\documentclass{article}

\usepackage{microtype}
\usepackage{graphicx}
\usepackage{subfigure}
\usepackage{booktabs} 

\usepackage{efbox,graphicx}
\efboxsetup{linecolor=black,linewidth=3pt}

\usepackage[T1]{fontenc}
\usepackage[normalem]{ulem}
\usepackage[utf8]{inputenc}
\usepackage{multirow}

\usepackage[accepted]{icml2020}

\usepackage[utf8]{inputenc}
\usepackage{amsmath,amsfonts,amssymb,amsthm,dsfont}

\usepackage{hyperref}
\usepackage{url}

\usepackage{todonotes}
\usepackage{float}

\def\Z{\mathbb{Z}}
\def\R{\mathbb{R}}

\def\1{\mathbbm{1}}

\def\P{\mathcal{P}}
\def\X{\mathcal{X}}
\def\Z{\mathcal{Z}}
\def\E{\mathcal{E}}
\def\D{\mathcal{D}}

\def\1{\mathds{1}}

\def\our{HyperCloud}

\icmltitlerunning{Hypernetwork approach to generating point clouds}

\begin{document}

\twocolumn[
\icmltitle{Hypernetwork approach to generating point clouds}

\icmlsetsymbol{equal}{*}

\begin{icmlauthorlist}

\icmlauthor{Przemysław Spurek}{to}
\icmlauthor{Sebastian Winczowski}{to}
\icmlauthor{Jacek Tabor}{to}
\icmlauthor{Maciej Zamorski}{goo}
\icmlauthor{Maciej Zięba}{goo}
\icmlauthor{Tomasz Trzciński}{ed}
\end{icmlauthorlist}
\icmlaffiliation{to}{Faculty of Mathematics and Computer Science, Jagiellonian University, Kraków, Poland}
\icmlaffiliation{goo}{Wrocław University of Science and Technology, Wrocław, Poland}
\icmlaffiliation{ed}{Warsaw University of Technology, Warsaw, Poland}

\icmlcorrespondingauthor{Przemysław Spurek}{przemyslaw.spurek@uj.edu.pl}

\icmlkeywords{Piont cloud, Generative models, mesh}

\vskip 0.3in
]

\printAffiliationsAndNotice{\icmlEqualContribution} 

\begin{abstract}
In this work, we propose a novel method for generating 3D point clouds that leverage properties of hyper networks. Contrary to the existing methods that learn only the representation of a 3D object, our approach simultaneously finds a representation of the object and its 3D surface.
The main idea of our \our{} method is to build a hyper network that returns weights of a particular neural network (target network) trained to map points from a uniform unit ball distribution into a 3D shape. As a consequence, a particular 3D shape can be generated using point-by-point sampling from the assumed prior distribution and transforming sampled points with the target network. Since the hyper network is based on an auto-encoder architecture trained to reconstruct realistic 3D shapes, the target network weights can be considered a parametrization of the surface of a 3D shape, and not a standard representation of point cloud usually returned by competitive approaches.
The proposed architecture allows finding mesh-based representation of 3D objects in a generative manner while providing point clouds en pair in quality with the state-of-the-art methods.
\end{abstract}

\section{Introduction}

Today many registration devices, such as LIDARs and depth cameras, are able to capture not only RGB channels, but also depth estimates. As a result, 3D objects registered by those devices and geometric data structures representing them, called point clouds, become increasingly important in contemporary computer vision applications, including autonomous driving~\cite{yang2018pixor} or robotic manipulation~\cite{kehoe2015survey}. To enable processing of point clouds, researchers typically transform them into regular 3D voxel grids or collections of images~\cite{su2015multi,wu20153d}. This, however, increases memory footprint of object representations and leads to significant information losses. On the other hand, representing 3D objects with the parameters of their surfaces is not trivial due to the complexity of mesh representations and combinatorial irregularities. Last but not least, point clouds can contain a variable number of data points corresponding to one object and registered at various angles, which requires for the methods that process them to be permutation and rotation invariant.

\begin{figure}
\begin{center} 
 \includegraphics[height=3.7cm]{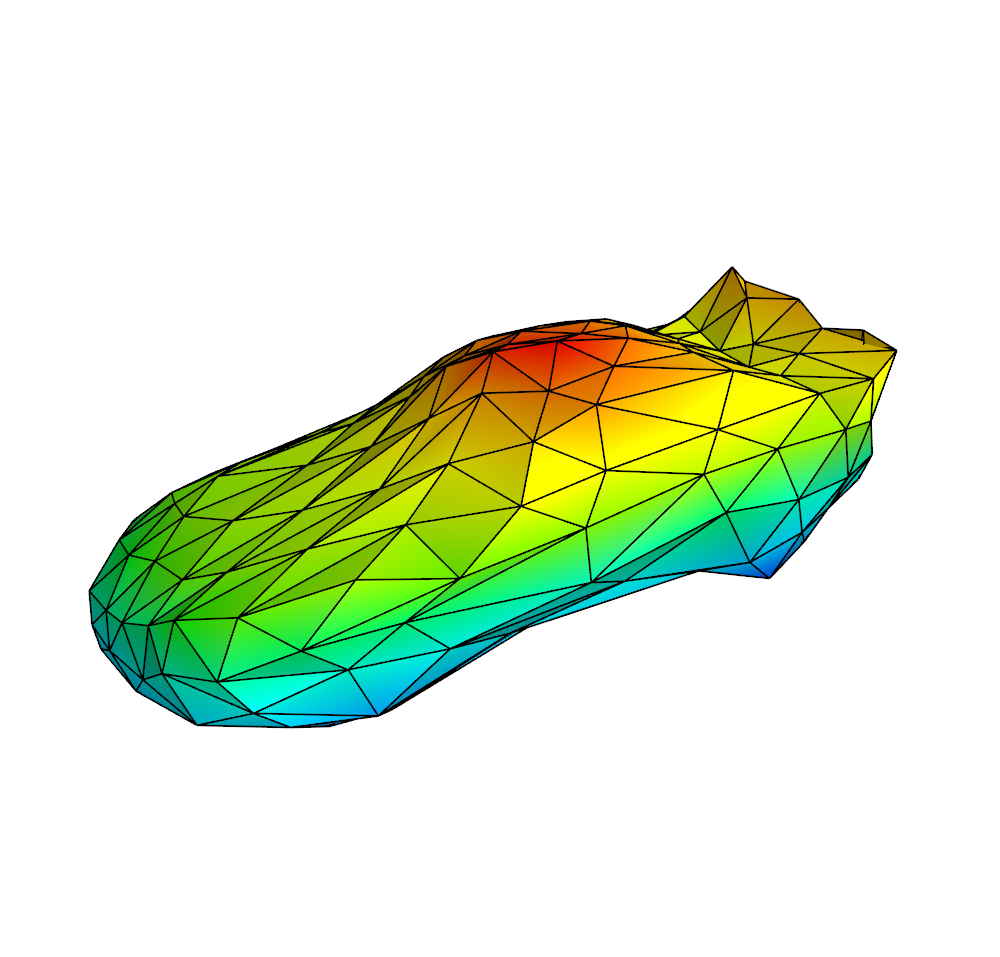}
 \includegraphics[height=3.5cm]{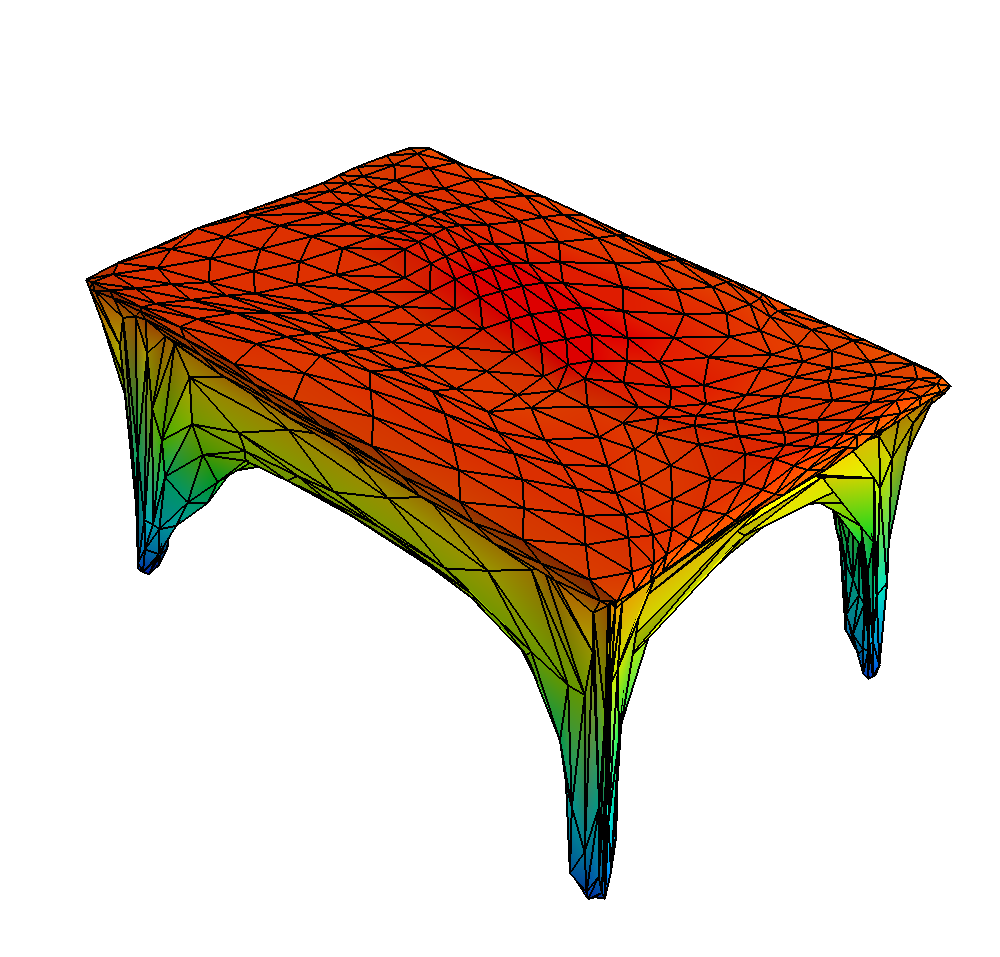}
 \includegraphics[height=3.7cm]{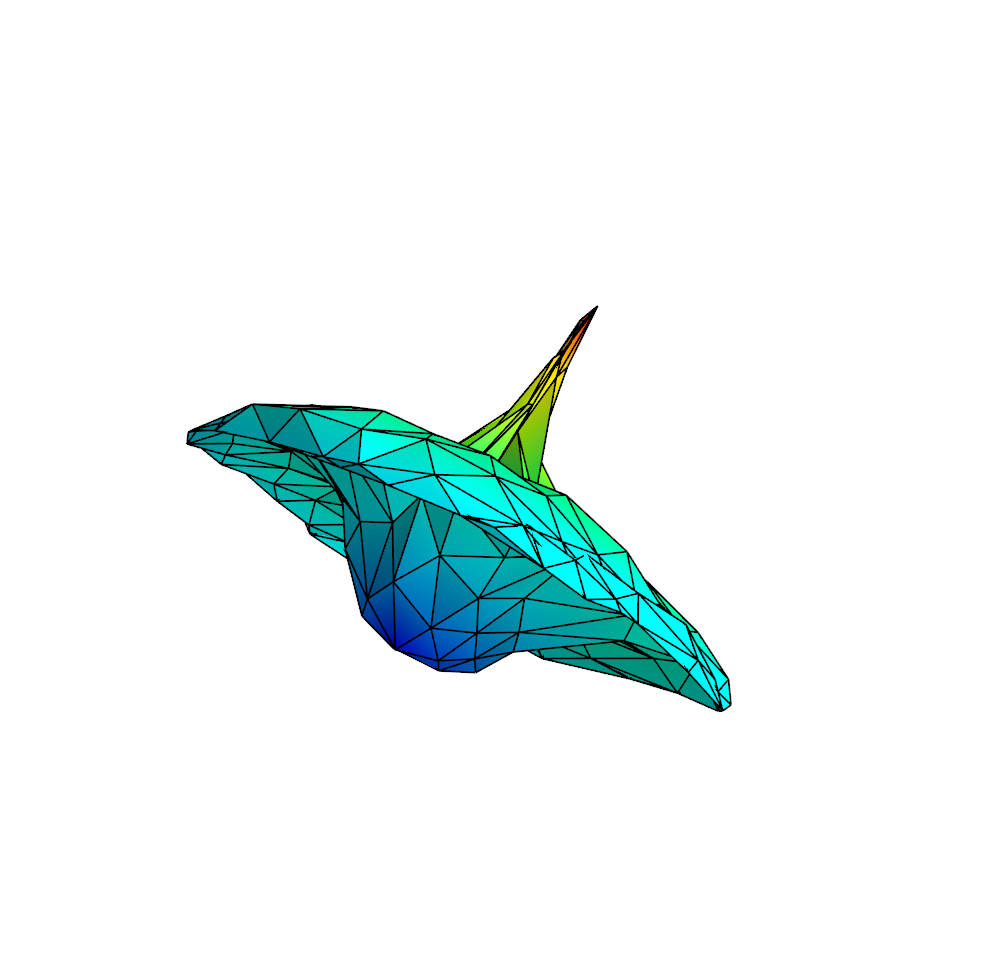}
 \includegraphics[height=3.7cm]{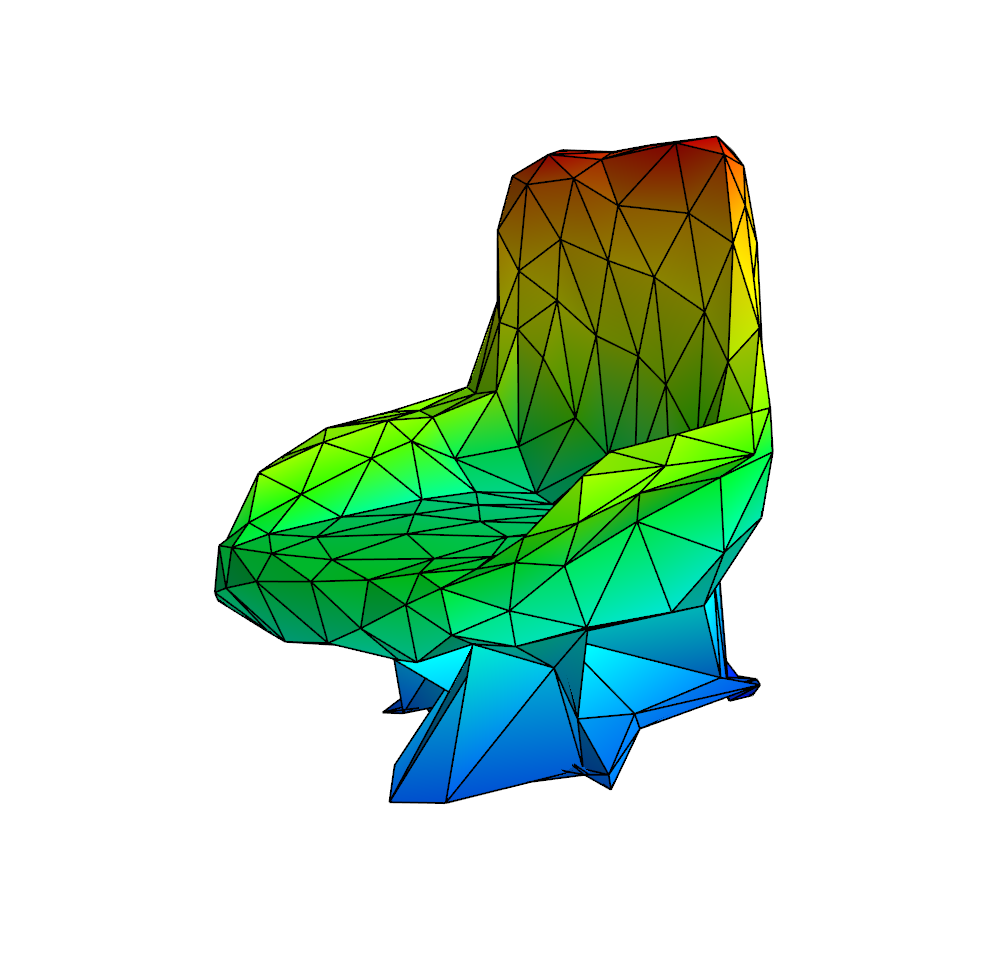}
\end{center} \vspace{-0.3cm}
\caption{Mesh representations generated by our \our{} method. Contrary to the existing methods that return point cloud representations sparsely distributed in 3D space, our approach allows to create a continuous 3D object representation in the form of high quality meshes.}
\vspace{-0.3cm}
\label{fig:ec_mesh} 
\end{figure} 

One way of addressing the above challenges related to point cloud representations is to subsample the point clouds and enforce permutation invariance within the model architecture, as it was done in DeepSets~\cite{zaheer2017deep} or PointNet~\cite{qi2017pointnet,qi2017pointnet++}. Although it works perfectly fine when point clouds are given as {\it an input} of the model, it is not obvious how to apply this approach for variable size {\it outputs}. Recently introduced family of methods solve for this problem by relying on generative models that return probability distribution of the points on the object surface, instead of an exact set of points~\cite{yang2019pointflow,stypulkowski2019conditional}. The most successful methods that follow this path, such as PointFlow~\cite{yang2019pointflow} and Conditioned Invertible Flow~\cite{stypulkowski2019conditional}, are based on the flow architecture that allows obtaining a representation of 3D object surfaces. The main limitation of the flow-based models is their cumbersome training process. Since flow architectures require the determinant of the Jacobian to be tractable for a given transformation, their optimization needs to be overly constrained.
Moreover, flow-based methods cannot be trained on probability distributions without compact support. 
For instance, it is not possible to train flow-based model on 3D ball since computing a cost function using log-likelihood returns infinity as a result and can therefore lead to numerical instability of the entire training procedure.
Moreover, flow-based models require the dimensionality of input and output data to be identical. Last but not least, these architectures need a significant amount of parameter and structure fine-tuning to work. 

\begin{figure}[!t] 
\begin{center} 
 \centering
 \includegraphics[height=3.0cm]{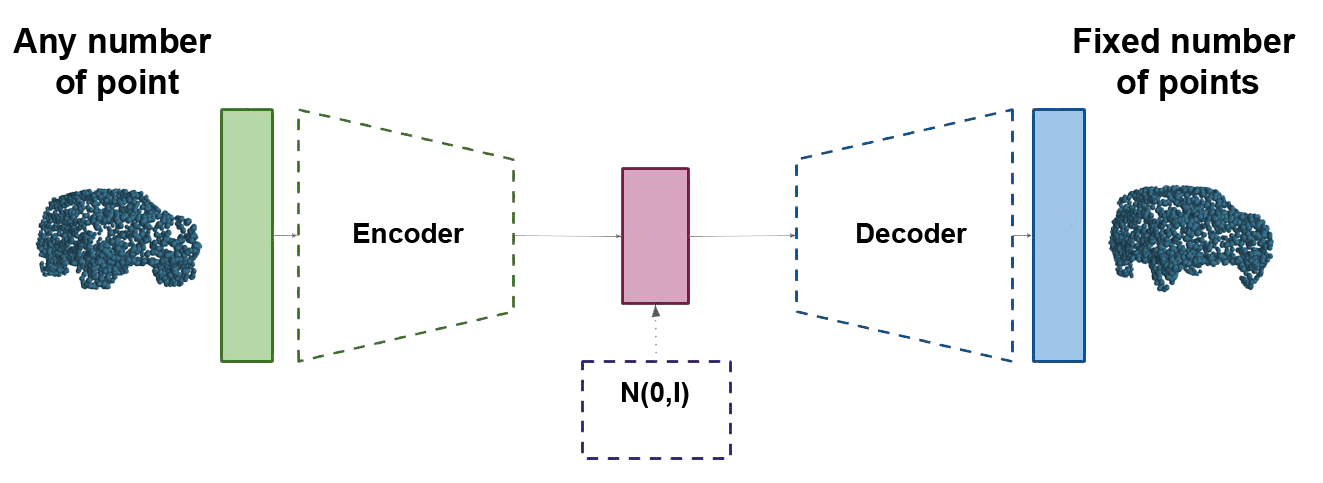} 
 \vspace{0.03cm}
 \hrule
 \vspace{0.3cm}
  \includegraphics[height=4.5cm]{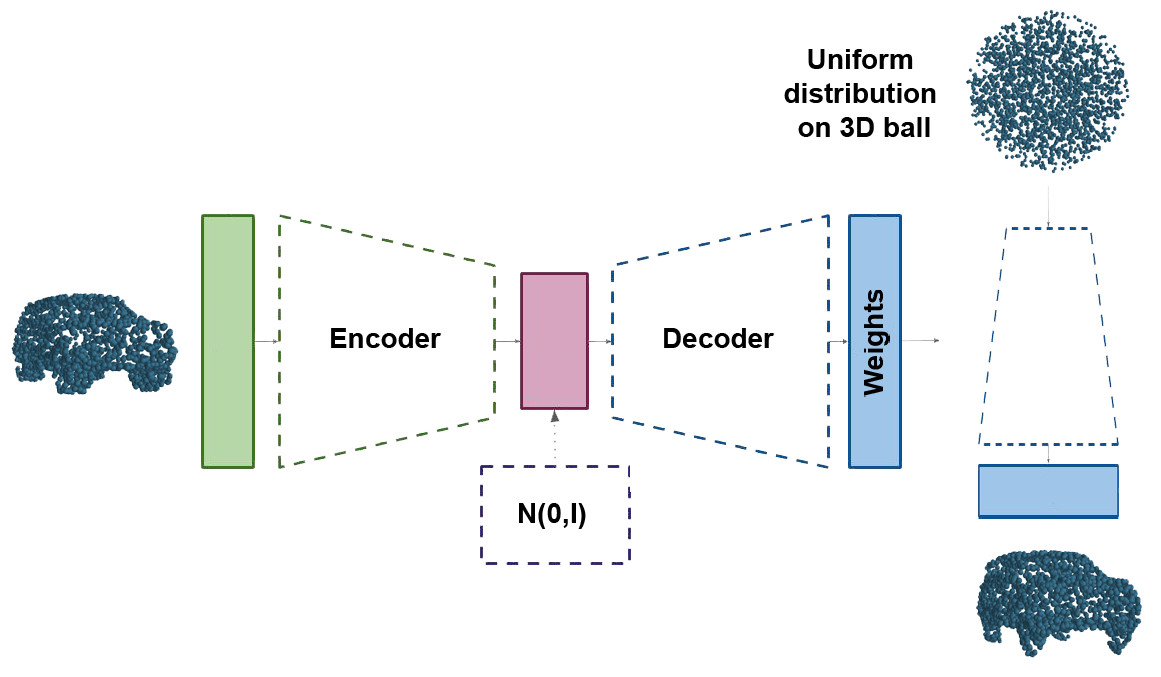} 
  \end{center}
  \caption{{\bf Top:} The baseline approach for generating 3D point clouds returns a fixed number of points~\cite{zamorski2018adversarial}. 
{\bf Bottom:} Our \our{} method leverages a hypernetwork architecture that takes a 3D point cloud as an input while returning the parameters of the {\it target network}. Since the parameters of the target network are generated by hypernetwork, the output dataset can be variable in size. As a result, we obtain a continuous parametrization of the object's surface and a more powerful representation of its mesh.}
 
\label{fig:teaser} 
\end{figure} 

\begin{figure}[!t] 
\begin{center} 
 \includegraphics[height=4.5cm]{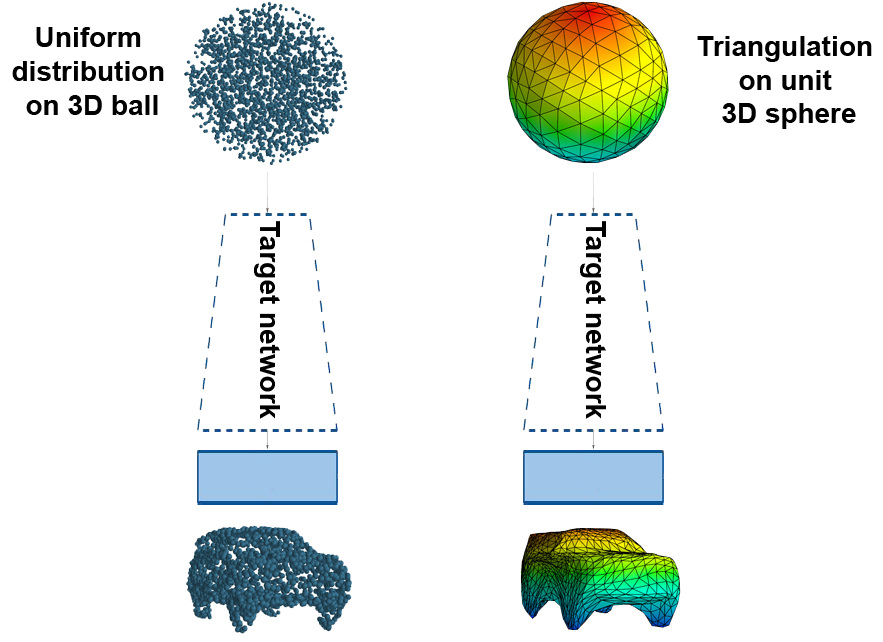} 
\end{center} 
  \caption{Scheme of producing mesh representations with {\our{}}. When using 3D {\emph{ball}} distribution, our method can generate 3D point clouds filled with data points, while when given 3D \emph{sphere} distribution it transforms samples from the sphere to surfaces of 3D objects - a feature highly desirable in the context of 3D mesh rendering.}
\label{fig:mesh} 
\end{figure}

In this paper, we propose to address the above shortcomings of the flow models by introducing a novel architecture that builds on the approach of~\cite{zamorski2018adversarial} and extends it with a hyper network~\cite{ha2016hypernetworks,klocek2019hypernetwork} that outputs weights of a generative model, the so-called {\it target network}. The target network can then be used to create an arbitrary number of points (depending on its architecture returned by a hyper network), instead of fixed-size sets. Fig.~\ref{fig:teaser} shows the overview of our method in comparison to the baseline approaches. Contrary to the flow-based models, our method dubbed {\bf \our{}}\footnote{We make our implementation available at \url{https://github.com/gmum/3d-point-clouds-HyperCloud}}  is much simpler conceptually and more general as it can be used to adapt any PointNet model to generate continuous output representation. Furthermore, it is much easier to train than the competing algorithms, as it requires a smaller number of hyperparameters and does not put any constraints on the input probability distribution and its Jacobian.
Finally, as presented in Fig.~\ref{fig:mesh}, our method returns a continuous mesh representation of 3D objects at virtually no cost in the quality of reconstructions. To the best of our knowledge, this is the first time a hyper network is used in the context of 3D point cloud generation, and we believe it opens a new research path into understanding and processing this type of data.

The contributions of this work can be summarized as follows: firstly, we introduce a novel yet general method that builds varied-size representations of point clouds that can be output by any model. Secondly, we achieve this by mapping the probability distributions to 3D models with generative target networks trained by a hypernetwork introduced in this work. Lastly, our approach offers a continuous mesh representation of 3D objects that can be used to render their surfaces directly, as shown in~Fig.~\ref{fig:ec_mesh}.

The remainder of this paper is structured as follows. Sec.~\ref{sec:related} discusses related works. In Sec.~\ref{sec:method} we introduce our \our{} approach and describe it in details. Sec.~\ref{sec:experiments} presents the results of evaluations and we conclude this work in Sec.~\ref{sec:conclusions}.

\section{Related work} 
\label{sec:related}

Introducing deep learning in the context of 3D point cloud representations allowed to improve performance in various discriminative tasks including classification \cite{qi2017pointnet,qi2017pointnet++,yang2018foldingnet,zaheer2017deep} and segmentation \cite{qi2017pointnet,shoef2019pointwise}.
Despite those successes, generating 3D point clouds with deep learning models remains a challenging task.

Due to the irregular format of point cloud representation, most researchers transform such data to regular 3D voxel grids or collections of images. In \cite{wu20153d}, the authors propose the voxelized representation of an input point cloud. Other approaches use multi-view 2D images \cite{su2015multi} or occupancy grid calculation \cite{ji20123d,maturana2015voxnet}.
Modeling volumetric objects in a general-adversarial manner is also considered in~\cite{wu2016learning} for the 3D-GAN model. 

Another approach to generative models for point cloud converts a point distribution to a $N \times 3$ matrix by sampling a pre-defined number of $N$ points from the distribution so that existing generative models are applicable. Such a solution can be applied in the VAE framework~\cite{gadelha2018multiresolution} as well as in adversarial auto-encoders (AAEs)~\cite{zamorski2018adversarial}. 

In the above methods, auto-encoders and GANs are trained with loss functions that optimize directly the distance between two point sets, e.g. using Chamfer distance (CD) or earth mover’s distance (EMD). In  \cite{sun2018pointgrow},  authors apply auto-regressive models \cite{van2016conditional} with a discrete point distribution to generate one point at a time, also using a fixed number of points per shape.

All the above methods learn to produce a fixed number of points for each shape, but they do not parametrize a surface of the shapes. Treating a point cloud as a fixed-dimensional matrix has several drawbacks. First, the model is restricted to generate a fixed number of points. Getting more points for a particular shape requires separate up-sampling models such as \cite{yifan2019patch,yu2018pu}. 

In \cite{yang2019pointflow}, authors propose a principled probabilistic framework to generate 3D point clouds by modeling them as a distribution
of distributions. 
PointFlow uses two-level of distributions where the first level is the distribution of shapes, and the second level is the distribution of points given a shape. PointFlow uses continuous normalizing flow \cite{chen2018neural, grathwohl2018ffjord} for both of these tasks.

Instead of directly parametrizing the
distribution of points in a shape, PointFlow models this distribution as an invertible parameterized transformation of 3D points from a prior distribution (e.g., a 3D Gaussian). Intuitively, under this model, generating points for a given shape involves sampling points from a generic Gaussian prior and then moving them according to this parameterized transformation to their new location in the target shape. Such solution has many advantages over the classical approaches, which only produce a cloud of points, nevertheless is is limited in multiple ways. The most important limitation is the fact that they use log-likelihood as a cost function, and, in consequence, cannot be trained on probability distributions with compact support. This significantly reduces the utility of flow-based models as, for instance, using a 3D ball distribution as a prior returns infinite values and therefore leads to numerical instability of training. In this work, we show that once this constraint is dropped thanks to using a fully-connected neural network we can directly model 3D point cloud surfaces and hence create their continuous mesh representations.

%

\section{\our{}: hypernetwork for generating 3D point clouds}
\label{sec:method}

In this section, we present our \our{} model for generating 3D point clouds. \our{} encompasses previously introduced approaches: the auto-encoder based generative model proposed in \cite{zamorski2018adversarial} and the hypernetwork proposed in \cite{ha2016hypernetworks}. Before we present our solution, we will briefly describe these two approaches.

\paragraph{Adversarial Auto-encoders for 3D Point Clouds}

Let us start with the auto-encoder architecture for 3D point cloud. Let 
$\X = \{X_i\}_{i=1,\ldots,n} = \{(x_i,y_i,z_i)\}_{i=1,\ldots,n}$  be a given dataset containing point clouds. The basic aim of auto-encoder is to transport the data through a typically, but not necessarily, lower dimensional latent space $\Z \subseteq \R^D$ while minimizing the reconstruction error.  Thus, we search for an encoder $\E:\X \to \Z$ and decoder $\D:\Z \to \X$ functions, which minimizes the reconstruction error between $X_i$ and its reconstructions $\D(\E X_i)$.

For point cloud representation, the crucial step is to define proper reconstruction loss that can be used in the autoencoding framework. In the literature, two common distance measures are successively applied for reconstruction purposes: Earth Mover’s (Wasserstein) Distance \cite{rubner2000earth} and Chamfer pseudo-distance \cite{tran20133d}.

Earth Mover’s Distance (EMD) is
a metric between two distributions based on the minimal
cost that must be paid to transform one distribution into the
other. For two equally sized subsets $X_1 \subset \R^3$ and $X_2 \subset \R^3$
their EMD is defined as:
$$
\begin{array}{c}
EMD(X_1,X_2)=\min\limits_{\phi:X_1\to X_2} \sum\limits_{x \in X_1} c(x, \phi(x))
\end{array}
$$

where $\phi$ is a bijection and $c(x, \phi(x))$ is the cost function and can be defined as: 
$$
\begin{array}{c}
c(x, \phi(x)) = \frac{1}{2} \| x - \phi(x) \|_2^2.
\end{array}
$$


Chamfer pseudo-distance (CD): measures the squared
distance between each point in one set to its nearest neighbor in the other set:
$$
\begin{array}{c}
CD(X_1,X_2)=\!\!\!\! \sum\limits_{x \in X_1} \min\limits_{y\in X_2} \! \| x-y \|\_2^2\! + \!\!\!\! \sum\limits_{x \in X_2} \min\limits_{y\in X_1} \! \| x-y \|_2^2.
\end{array}
$$

Auto-encoder based generative model is a classical auto-encoder model with a modified cost function, which forces the model to be generative, i.e., ensures that the data transported to the latent space comes from the prior distribution (typically Gaussian). 
Thus, to construct a generative auto-encoder model, we add to its cost function a measure of the distance of a given sample from prior distribution.

Variational Auto-encoderss (VAE) are generative models that are capable of learning approximated data distribution by applying variational inference \cite{kingma2013auto}.
To ensure that the data transported to latent space $\Z$ are distributed according to standard normal density, we add the distance from standard multivariate normal density:

\begin{figure*}[!h] 
\begin{center}  
 \includegraphics[height=14cm]{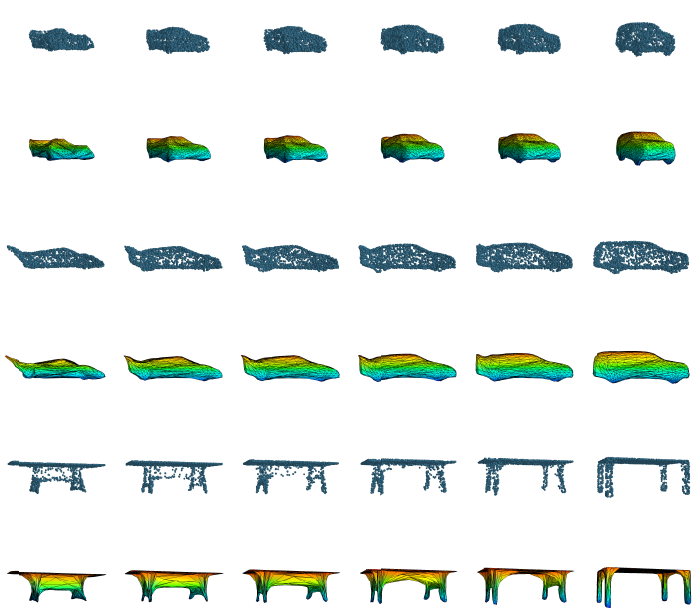}
\end{center} 
  \caption{Interpolations between two 3D point clouds and its mesh representations.} 
\label{fig:inter_obj} 
\end{figure*} 

$$
\begin{array}{c}
\mathrm{cost}(\X;\E,\D) \! =  \! Err(\X;\D(\E \X )) \!\! + \!\! \lambda D_{KL}(\E \X,N(0,I)),
\end{array}
$$
where $ D_{KL}$ is the Kullback–Leibler divergence \cite{kullback1951information}.

The main limitation of VAE models is that the regularization term requires particular prior distribution to make KL
divergence tractable. In order to deal with that limitation, the authors of \cite{makhzani2015adversarial} introduce Adversarial Auto-encoder (AAE) that utilize adversarial training to force a particular distribution on $\Z$ space. The model assumes that an additional neural network - discriminator, which is responsible for
distinguishing between fake and true samples, where the
true samples are sampled from an assumed prior distribution
and fake samples are generated via an encoding network.

In \cite{zamorski2018adversarial}, authors propose an approach to Adversarial Auto-encoders dedicated to the 3D point clouds.
Because the input of the model is a set of points, they use as encoder $\E$ PointNet model \cite{qi2017pointnet} that is invariant to permutations. We receive the same distribution for all possible orderings of points from $X$. Since discriminator is not permutation invariant mapping $D$ (as it is a simple MLP model), authors utilize an additional function that provides one-to-one mapping for the points stored in $X$. 

The probability distribution assumed on latent space can be more complex than $N(0,I)$ and not given in an explicit form. Some autoencoders try to learn some more sophisticated distributions directly from data. Such solutions may utilize techniques like VampPrior \cite{tomczak2017vae} or incorporate continuous \cite{yang2019pointflow} or discrete \cite{berg2018sylvester} normalizing flows.   

Due to large techniques of enforcing probability distribution on the latent space, the cost function of the model can be formulated in the more general form:
\begin{equation}
\mathrm{cost}(\X;\E,\D)= Err(\X;\D(\E \X )) + Reg(\E \X, P),
\label{eq:cost_general}
\end{equation}
where $Err$ is Earth Mover’s (Wasserstein) Distance or Chamfer pseudo-distance and $Reg$ is a function that forces latent space to be from some known or trainable distribution $P$. For known distributions like Gaussian, Kullback–Leibler divergence or adversarial training can be used for regularization. 



In our work, we propose to enrich the presented regularized autoencoder by replacing the decoder with the hypernetwork. The goal of the hypernetwork is to transform the latent representation of the point cloud to the weights of the so-called target network. The goal of the target network is to transform the samples from assumed prior to the points that represent 3D shapes without assuming the arbitrary number of points. Roughly speaking, in our case, hypernetwork produces a parametrization of the respective generative model.

\paragraph{Hyper-network}

Hyper-networks, introduced in \cite{ha2016hypernetworks}, are defined as neural models that generate weights for a separate target network solving a specific task. 
The authors aim to reduce the number of trainable parameters by designing a hyper-network with a smaller number of parameters than the target network. Making an analogy between hyper-networks and generative models, the authors of \cite{sheikh2017stochastic}, use this
mechanism to generate a diverse set of target networks approximating the same function. 


\begin{figure*}[!h] 
\begin{center} 
 \includegraphics[height=5.5cm]{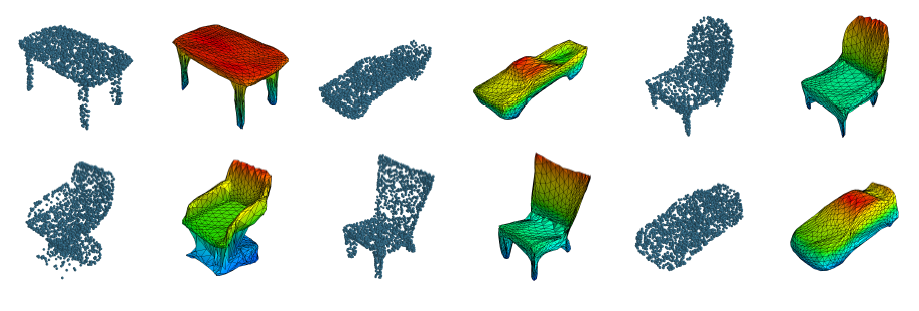}
\end{center} 
  \caption{3D point clouds and their mesh representations produced by \our{}.} 
\label{fig:generated_point} 
\end{figure*}

Hyper-networks can also be used for functional representations of images \cite{klocek2019hypernetwork}. In such concept by a functional (or deep) representation of an image, authors understand a function (neural network) $I : \R^2 \to \R^3$ which given a point (with arbitrary coordinates) $(x, y)$ in the plane returns the point in $[0, 1]^3$ representing the RGB values of the color of the image at $(x, y)$.

\paragraph{\our{}} Inspired by the above methods, we propose our {\our{}} model that uses hyper-network to output weights of generative network to create 3D point clouds, instead of generating them directly with the decoder, as done in~\cite{zamorski2018adversarial}. More specifically, we present parameterization of the surface of 3D objects as a function $S : \R^3 \to \R^3$, which given a point from the prior distribution $(x, y, z)$ returns the point on the surface of the objects. 
Roughly speaking, instead of producing a 3D point cloud, we would like to produce many neural networks (a different neural network for each object) that model surfaces of objects. 


In practice, we have one neural network architecture that uses different weights for each 3D object.
More precisely, we model function  $T_{\theta} : \R^3 \to \R^3$ (neural network with weights $\theta$), which takes an element from the prior distribution $\P$ and transfers it on an element on the surface of the object. In our work, we use the transformation between uniform distribution on the 3D ball and the object. 
This choice of distribution allows one to create a continuous mesh representation.
The key idea behind this is that the distribution does not have compact support.
Roughly speaking, Gaussian distribution does not have a smooth border.

In consequence, we can produce as many points as we need (we can sample an arbitrary number of points from the uniform distribution of the unit ball and transfer them by target network). Thanks to the target network, we can train our model on point clouds containing a different number of points. 

Furthermore, we can produce a  continuous mesh representation of the object.
All elements from the ball are transformed into a 3D object. In consequence, the unit sphere is transformed into the border of our data set.
Now we can produce meshes without a secondary mesh rendering procedure. It is obtained by simply feeding our neural network by the vertices of a sphere mesh, see Fig~\ref{fig:mesh}. As a result, we obtain a high-quality meshes of 3D objects. The sharpness of the borders is a direct consequence of compact support probability distribution of the input prior. Since flow-based models cannot handle this family of priors and require infinite support distributions, the representations generated with those models are lower quality.

The target network is not trained directly.   
We use a hyper-network
$
\begin{array}{c}
H_{\phi}: \R^3 \supset X \to \theta ,
\end{array}
$
which for an point-cloud $X \subset \R^3$ returns weights $\theta$ to the corresponding target network $T_{\theta}$.
Thus, a point cloud $X$ is represented by a function 
$$
\begin{array}{c}
T((x,y,x);\theta) = T((x,y,x); H_{\phi}(X)).
\end{array}
$$


To use the above model, we need to train the weights $\phi$ of the hypernetwork. For this purpose, we minimize the distance between point clouds like Chamfer distance (CD) or earth mover’s distance (EMD) over the training set of points clouds. More precisely, we take an input point cloud $X \subset \R^3$ and pass it to $H_{\phi}$. The hypernetwork returns weights $\theta$ to target network $T_{\theta}$. Next, the input point cloud $X$ is compared with the output from the target network $T_{\theta}$ (we sample the correct number of points from the prior distribution and transfer them by target network). 
As a hypernetwork, we use a permutation invariant encoder that is based on PointNet architecture \cite{qi2017pointnet} and modified decoder to produce weight instead of row points.
The architecture of $T_{\theta}$ consists of: an encoder ($\E$) which is a PointNet-like network that transports the data to lower-dimensional latent space $\Z \in \R^D$ and a decoder ($\D$)  (fully-connected
network), which transfers latent space to the vector of weights for the target network. 
In our framework hypernetwork $T_{\theta}(X)$ represents our autoencoder structure $\D(\E X)$.
Assuming  $T_{\theta}(X) = \D(\E X)$, we train our model by minimizing the cost function given by equation (\ref{eq:cost_general}). 



Observe, that we only train a single neural model (hypernetwork), which allows us to produce a great variety of functions at test time. In consequence, we might expect that target networks for similar point cloud will be similar (see  Sec. \ref{sec:experiments} for details). We are able to produce smooth interpolation by using hypernetwork.

\begin{figure}[!h] 
\begin{center} 
 \includegraphics[height=3.2cm]{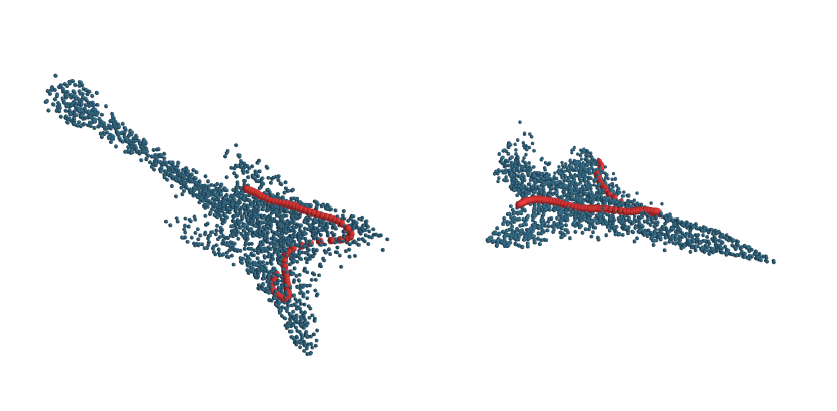}  
\end{center} 
  \caption{
  Thanks to using hypernetwork architecture, we can work with one object (distribution of points on a single 3D point cloud). One possible application is interpolation in the target network instead. By taking two samples on the uniform ball and its interpolation we can construct interpolation between points on the surface of the object.} 
\label{fig:inter_point} 
\end{figure}

\begin{table*}[]
\begin{center}
\scalebox{0.85}{
\begin{tabular}{llccccccc}
\hline
\multirow{2}{*}{Cathegory} & \multirow{2}{*}{Methods} & \multirow{2}{*}{JSD} & \multicolumn{2}{c}{MMD} & \multicolumn{2}{c}{COV} & \multicolumn{2}{c}{1-NNA} \\ \cline{4-9} 
                           &                          &                      & CD          & EMD       & CD         & EMD        & CD          & EMD         \\ \hline
\multirow{7}{*}{Airplane}  & r-GAN                    & 7.44                 & 0.261       & 5.47      & 42.72      & 18.02      & 93.58       & 99.51       \\
                           & l-GAN (CD)               & 4.62                 & 0.239       & 4.27      & 43.21      & 21.23      & 86.30       & 97.28       \\
                           & l-GAN (EMD)              & \bf 3.61                 & 0.269       & 3.29      & \bf 47.90      & \bf 50.62      & 87.65       & 85.68       \\
                           & PC-GAN                   & 4.63                 & 0.287       & 3.57      & 36.46      & 40.94      & 94.35       & 92.32       \\
                           & PointFlow                & 4.92                 & \bf 0.217       & \bf 3.24      & 46.91      & 48.40      & \bf 75.68       & \bf 75.06       \\
                           & \our{} (ours)      & 4.84                 & 0.266       & 3.28      & 39.75      & 43.70      & 93.80       & 88.95       \\ \cline{2-9} 
                           & Training set             & 6.61                 & 0.226       & 3.08      & 42.72      & 49.14      & 70.62       & 67.53       \\ \hline
\multirow{7}{*}{Chair}     & r-GAN                    & 11.5                 & 2.57        & 12.8      & 33.99      & 9.97       & 71.75       & 99.47       \\
                           & l-GAN (CD)               & 4.59                 & 2.46        & 8.91      & 41.39      & 25.68      & 64.43       & 85.27       \\
                           & l-GAN (EMD)              & 2.27                 & 2.61        &  7.85      & 40.79      & 41.69      & 64.73       & 65.56       \\
                           & PC-GAN                   & 3.90                 & 2.75        & 8.20      & 36.50      & 38.98      & 76.03       & 78.37       \\
                           & PointFlow                & \bf 1.74                 & \bf 2.42        & 7.87      & \bf 46.83      & \bf 46.98      & \bf 60.88       & \bf 59.89       \\
                           & \our{} (ours)      & 2.73                 & 2.56        & \bf 7.84      & 41.54      & 46.67      & 68.20       & 68.80       \\ \cline{2-9} 
                           & Training set             & 1.50                 & 1.92        & 7.38      & 57.25      & 55.44      & 59.67       & 58.46       \\ \hline
\multirow{7}{*}{Car}       & r-GAN                    & 12.8                 & 1.27        & 8.74      & 15.06      & 9.38       & 97.87       & 99.86       \\
                           & l-GAN (CD)               & 4.43                 & 1.55        & 6.25      & 38.64      & 18.47      & 63.07       & 88.07       \\
                           & l-GAN (EMD)              & 2.21                 & 1.48        & 5.43      & 39.20      & 39.77      & 69.74       & 68.32       \\
                           & PC-GAN                   & 5.85                 & 1.12        & 5.83      & 23.56      & 30.29      & 92.19       & 90.87       \\
                           & PointFlow                & \bf 0.87                 & \bf 0.91        & \bf 5.22      & \bf 44.03      & \bf 46.59      & \bf 60.65       & \bf 62.36       \\
                           & \our{} (ours)      & 3.09                 & 1.07        & 5.38      & 40.05      & 40.05      & 84.65       & 77.27       \\ \cline{2-9} 
                           & Train set                & 0.86                 & 1.03        & 5.33      & 48.30      & 51.42      & 57.39       & 53.27       \\ \hline
\end{tabular}
}
\caption{Generation results. MMD-CD scores are multiplied by
$10^3$; MMD-EMD scores and JSDs are multiplied by $10^2$. }
\label{tab:gen_results}
\end{center}
\end{table*}


\section{Experiments}
\label{sec:experiments}

In this section, we describe the experimental results of the proposed generative models in various tasks, including 3D points mesh generation and interpolation. In the first subsection, we show that our model inherited reconstruction and generative capabilities from models based on generating a fixed number of points. Then we show that we are able to produce continuous mesh representation.

\paragraph{Metrics}





Following the methodology for evaluating generative fidelity and diversification among samples provided in \cite{achlioptas2017learning} and \cite{yang2019pointflow}, we utilize the following criteria for evaluation: Jensen-Shannon Divergence, Coverage, Minimum Matching Distance 1-nearest Neighbor Accuracy.

Jensen-Shannon Divergence (JSD): a measure of the distance between two empirical distributions $P$ and $Q$, defined as:
$$
\begin{array}{c}
JSD(P\|Q) \! = \! \frac{KL(P\|M) + KL(Q\|M)}{2}, \mbox{ where } M \! = \! \frac{P+Q}{2}.
\end{array}
$$

Coverage (COV): a measure of generative capabilities in terms of richness of generated samples from the model. For two point cloud sets $X_1, X_2 \subset \R$ coverage is defined as a fraction of points in $X_2$ that are in the given metric the nearest neighbor to some points in $X_1$.

Minimum Matching Distance (MMD): Since COV only takes the closest point clouds into account and does not depend on the distance between the matchings additional metric was introduced. For point cloud sets $X_1$, $X_2$ MMD is a measure of similarity between point clouds in $X_1$ to those in $X_2$.

1-Nearest Neighbor Accuracy (1-NNA) is a testing procedure characteristic for evaluating GANs. We consider two sets: set $S_g$ composed of generated point clouds and set of test (reference) point clouds, $S_r$. We pick some generated point cloud $X$ from $S_g$ and find the corresponding nearest neighbor in $S_{-X}=S_r \bigcup S_g - \{X\}$, the set that aggregates both training and sampled shapes excluding the considered point cloud $X$. The 1-NNA is the leave-one-out accuracy of the 1-NN classifier:
$$
\begin{array}{c}
1-NNA = \frac{\sum_{X \in S_g}\1 [N_X \in S_g] + \sum_{Y \in S_r} \1 [N_Y \in S_r]}{|S_g| + |S_r|}.    
\end{array}
$$

For each sample, the 1-NN classifier classifies it as coming from $S_r$ or $S_g$ according to the label of its nearest sample. The perfect situation occurs when the classifier is unable to distinguish between real and generated point clouds, which means that the value of the criterion is close to $50\%$.


\begin{table}[]
\scalebox{0.90}{
\begin{tabular}{clllll}
\hline
\multicolumn{1}{c|}{\multirow{2}{*}{Sphere R}} & \multicolumn{1}{c}{\multirow{2}{*}{JSD}} & \multicolumn{2}{c}{MMD}                                               & \multicolumn{2}{c}{COV}                                                 \\ \cline{3-6} 
\multicolumn{1}{c|}{}                          & \multicolumn{1}{c}{}                     & \multicolumn{1}{c}{CD}            & \multicolumn{1}{c}{EMD}           & \multicolumn{1}{c}{CD}             & \multicolumn{1}{c}{EMD}            \\ \hline
                                               & \multicolumn{5}{c}{\textit{Airplane}}                                                                                                                                                      \\ \hline
\multicolumn{1}{c|}{PointFlow}                 &                                          &                                   &                                   &                                    &                                    \\ \hline
\multicolumn{1}{c|}{R=2.795}                   & 22.26                                    & 0.49                              & 6.65                              & \textbf{44.69}                     & 20.74                              \\
\multicolumn{1}{c|}{R=3.136}                   & 26.46                                    & 0.60                              & 6.89                              & 39.50                              & 19.01                              \\
\multicolumn{1}{c|}{R=3.368}                   & 29.65                                    & 0.68                              & 6.84                              & 40.49                              & 16.79                              \\ \hline
\multicolumn{1}{c|}{\our{} (ours)}              &                                          &                                   &                                   &                                    &                                    \\ \hline
\multicolumn{1}{c|}{R=1}                       & \textbf{9.51}                            & \textbf{0.45}                     & \textbf{5.29}                     & 30.60                              & \textbf{28.88}                     \\ \hline
                                               & \multicolumn{5}{c}{\textit{Chair}}                                                                                                                                                         \\ \hline
\multicolumn{1}{c|}{PointFlow}                 &                                          &                                   &                                   &                                    &                                    \\ \hline
\multicolumn{1}{c|}{R=2.795}                   & \multicolumn{1}{c}{19.28}                & \multicolumn{1}{c}{4.28}          & \multicolumn{1}{c}{13.38}         & \multicolumn{1}{c}{36.85}          & \multicolumn{1}{c}{20.84}          \\
\multicolumn{1}{c|}{R=3.136}                   & \multicolumn{1}{c}{22.52}                & \multicolumn{1}{c}{4.89}          & \multicolumn{1}{c}{14.47}         & \multicolumn{1}{c}{32.47}          & \multicolumn{1}{c}{17.22}          \\
\multicolumn{1}{c|}{R=3.368}                   & 24.68                                    & 5.36                              & 14.97                             & 31.41                              & 17.06                              \\ \hline
\multicolumn{1}{c|}{\our{} (ours)}              &                                          &                                   &                                   &                                    &                                    \\ \hline
\multicolumn{1}{c|}{R=1}                       & \multicolumn{1}{c}{\textbf{4.32}}        & \multicolumn{1}{c}{\textbf{2.81}} & \multicolumn{1}{c}{\textbf{9.32}} & \multicolumn{1}{c}{\textbf{40.33}} & \multicolumn{1}{c}{\textbf{40.63}} \\ \hline
                                               & \multicolumn{5}{c}{\textit{Car}}                                                                                                                                                           \\ \hline
\multicolumn{1}{c|}{PointFlow }                 &                                          &                                   &                                   &                                    &                                    \\ \hline
\multicolumn{1}{c|}{R=2.795}                   & 16.59                                    & 1.6                               & 8.00                              & 20.17                              & 17.04                              \\
\multicolumn{1}{c|}{R=3.136}                   & 20.21                                    & 1.75                              & 7.80                              & 21.59                              & 17.32                              \\
\multicolumn{1}{c|}{R=3.368}                   & 24.10                                    & 1.96                              & 8.35                              & 18.75                              & 17.04                              \\ \hline
\multicolumn{1}{c|}{\our{} (ours)}              &                                          &                                   &                                   &                                    &                                    \\ \hline
\multicolumn{1}{c|}{R=1}                       & \textbf{5.20}                            & \textbf{1.11}                     & \textbf{6.54}                     & \textbf{37.21}                     & \textbf{28.40}                     \\ \hline
\end{tabular}
}
\caption{The values of quality measures of 3D representations obtained by sampling from sphere of a given radius $R$ for \emph{airplane}, \emph{chair} and \emph{car} shapes. It can be seen that \our{} preserves the good quality of sampled point clouds, while PointFlow has difficulties in obtaining good quality representations from the sphere. }
\label{tab:sphere}
\end{table}

We examine the generative capabilities of the provided \our{} model in comparison to the existing reference approaches. In this experiment, we follow the methodology provided in \cite{yang2019pointflow}. For this particular experiment, we utilize the hypernetwork architecture trained with EMD reconstruction loss together with the continuous flow on latent representation instead of simple KLD regularization. We compare the results with the existing solutions: raw-GAN \cite{achlioptas2017learning}, latent-GAN \cite{achlioptas2017learning}, PC-GAN \cite{li2018point} and PointFlow \cite{yang2019pointflow}. We train each model using point clouds from one of the three categories in the ShapeNet dataset: \emph{airplane},
\emph{chair}, and \emph{car}. We follow the exact evaluation pipeline provided in \cite{yang2019pointflow}. 

The results are presented in Table \ref{tab:gen_results}. The HyperNetwork obtains comparable results to the other models that utilize EMD reconstruction loss with the advantage of sampling an arbitrary number of points. The model was outperformed by PointFlow that does not utilize EMD as reconstruction loss and is not directly capable of generating 3D meshes. 


\paragraph{Generation of 3D meshes}

The main advantage of our method comparing to reference solutions is the ability to generate both 3D point clouds and meshes without any post-processing stage. 
In Fig.~\ref{fig:generated_point}, we present point cloud as well as mesh representation generated by the same model.
Thanks to using a uniform distribution on the 3D ball, we can easily construct mesh. All elements from the ball are transformed into a 3D object. In consequence, the unit sphere is transformed into the border of our data set. As it was mentioned, we can produce meshes without a secondary meshing procedure. It is obtained by propagating the triangulation of the 3D sphere through the target network, see Fig~\ref{fig:mesh}.

In the case of Gaussian prior, we can use a similar procedure, but it is nontrivial to select the optimal sphere radius, which will be used by the generation of mesh (contrary to HyperCloud, in PointFlow there is no default for radius $R$). If the chosen radius is too small, the constructed mesh lies {\em inside} the point cloud, and consequently, we lose small outlying elements of the object, e.g., chair legs.
On the other hand, if the chosen sphere radius is large, some small elements of the 3D object will be merged, e.g., four legs of a chair will be joined into one.

For evaluation of the quality of mesh grid representation, we propose the following experiment. Instead of sampling the points from the assumed prior distribution, we sample them from a given surface (sphere of the assumed radius). Next, we calculate the standard quality measures of generated point clouds considered in the previous experiment. Since all models except PointFlow listed in Tab.~~\ref{tab:gen_results} work only on a fixed number of points we compare our results only with PointFlow. 


As it was mention above, we can use the PointFlow model to produce mesh representation in a similar way by feeding the target network by triangulation on a sphere. In our experiment, consistently with the standard used for hypothesis testing, we use $95\%$, $98\%$ and $99\%$ confidence spheres for 3D Gaussian distribution, see Tab.~\ref{tab:sphere}. As we can see, the default Gaussian prior is not suitable for producing a continuous representation of the boundary. Moreover, the seemingly natural exchange (with accordance with our approach) of the normal distribution onto the uniform distribution on the ball will not work since flow methods use log-likelihood as a cost function, and consequently, it is impossible to use prior density with compact support. 



\paragraph{Interpolation}
In our model, we can construct two types of interpolation.
Since we have two different prior distributions: Gaussian in hyper network architecture (latent of auto-encoder) and uniform distribution on the unit sphere in the target network, see Fig.~\ref{fig:teaser}.
First of all, we can take two 3d objects and obtain a smooth transition between them, see Fig.~\ref{fig:inter_obj}. For each point cloud, we can generate mesh representation. Therefore we can also produce interpolation between meshes.  

Thanks to using hypernetwork architecture, we can work with one object (distribution of points on a single 3D point cloud).
One possible application is interpolation in the target network instead of the classical approach in the latent space of auto-encoder, see Fig.~\ref{fig:inter_point}. By taking two samples on the uniform ball and its interpolation, we can construct interpolation between points on the surface of the object.

\section{Conclusions}
\label{sec:conclusions}

In this work, we presented a novel approach to represent point clouds of 3D objects with parameters of target networks trained by a hypernetwork as generative models. More specifically, we are able to build variable size representations of point clouds not only when they are inputted into the model, but also when they are returned as an output. Contrary to the existing methods, our approach is not constrained by the assumptions enforced on the objective functions in the case of the flow-based architectures, such as tractability of Jacobian determinants. Finally, our {\our} method offers a general framework that allows to adapt any PointNet model to build a continuous representation of the output vector. In this work we focused specifically on mesh representations of 3D objects, presenting that our approach give empirically better results on the task of realistic mesh generation. Nevertheless, thanks to the generality of our proposed architecture that encompasses many existing ones, it can be used in a multitude of real-life applications and it can open new areas of research related to generative models.

\bibliographystyle{icml2020}

\begin{thebibliography}{34}
\providecommand{\natexlab}[1]{#1}
\providecommand{\url}[1]{\texttt{#1}}
\expandafter\ifx\csname urlstyle\endcsname\relax
  \providecommand{\doi}[1]{doi: #1}\else
  \providecommand{\doi}{doi: \begingroup \urlstyle{rm}\Url}\fi

\bibitem[Achlioptas et~al.(2017)Achlioptas, Diamanti, Mitliagkas, and
  Guibas]{achlioptas2017learning}
Achlioptas, P., Diamanti, O., Mitliagkas, I., and Guibas, L.
\newblock Learning representations and generative models for 3d point clouds.
\newblock \emph{arXiv preprint arXiv:1707.02392}, 2017.

\bibitem[Berg et~al.(2018)Berg, Hasenclever, Tomczak, and
  Welling]{berg2018sylvester}
Berg, R. v.~d., Hasenclever, L., Tomczak, J.~M., and Welling, M.
\newblock Sylvester normalizing flows for variational inference.
\newblock \emph{arXiv preprint arXiv:1803.05649}, 2018.

\bibitem[Chen et~al.(2018)Chen, Rubanova, Bettencourt, and
  Duvenaud]{chen2018neural}
Chen, T.~Q., Rubanova, Y., Bettencourt, J., and Duvenaud, D.~K.
\newblock Neural ordinary differential equations.
\newblock In \emph{Advances in neural information processing systems}, pp.\
  6571--6583, 2018.

\bibitem[Gadelha et~al.(2018)Gadelha, Wang, and
  Maji]{gadelha2018multiresolution}
Gadelha, M., Wang, R., and Maji, S.
\newblock Multiresolution tree networks for 3d point cloud processing.
\newblock In \emph{Proceedings of the European Conference on Computer Vision
  (ECCV)}, pp.\  103--118, 2018.

\bibitem[Grathwohl et~al.(2018)Grathwohl, Chen, Betterncourt, Sutskever, and
  Duvenaud]{grathwohl2018ffjord}
Grathwohl, W., Chen, R.~T., Betterncourt, J., Sutskever, I., and Duvenaud, D.
\newblock Ffjord: Free-form continuous dynamics for scalable reversible
  generative models.
\newblock \emph{arXiv preprint arXiv:1810.01367}, 2018.

\bibitem[Ha et~al.(2016)Ha, Dai, and Le]{ha2016hypernetworks}
Ha, D., Dai, A., and Le, Q.~V.
\newblock Hypernetworks.
\newblock \emph{arXiv preprint arXiv:1609.09106}, 2016.

\bibitem[Ji et~al.(2012)Ji, Xu, Yang, and Yu]{ji20123d}
Ji, S., Xu, W., Yang, M., and Yu, K.
\newblock 3d convolutional neural networks for human action recognition.
\newblock \emph{IEEE transactions on pattern analysis and machine
  intelligence}, 35\penalty0 (1):\penalty0 221--231, 2012.

\bibitem[Kehoe et~al.(2015)Kehoe, Patil, Abbeel, and Goldberg]{kehoe2015survey}
Kehoe, B., Patil, S., Abbeel, P., and Goldberg, K.
\newblock A survey of research on cloud robotics and automation.
\newblock \emph{IEEE Transactions on automation science and engineering},
  12\penalty0 (2):\penalty0 398--409, 2015.

\bibitem[Kingma \& Welling(2013)Kingma and Welling]{kingma2013auto}
Kingma, D.~P. and Welling, M.
\newblock Auto-encoding variational bayes.
\newblock \emph{arXiv preprint arXiv:1312.6114}, 2013.

\bibitem[Klocek et~al.(2019)Klocek, Maziarka, Wo{\l}czyk, Tabor, Nowak, and
  {\'S}mieja]{klocek2019hypernetwork}
Klocek, S., Maziarka, {\L}., Wo{\l}czyk, M., Tabor, J., Nowak, J., and
  {\'S}mieja, M.
\newblock Hypernetwork functional image representation.
\newblock In \emph{International Conference on Artificial Neural Networks},
  pp.\  496--510. Springer, 2019.

\bibitem[Kullback \& Leibler(1951)Kullback and
  Leibler]{kullback1951information}
Kullback, S. and Leibler, R.~A.
\newblock On information and sufficiency.
\newblock \emph{The annals of mathematical statistics}, 22\penalty0
  (1):\penalty0 79--86, 1951.

\bibitem[Li et~al.(2018)Li, Zaheer, Zhang, Poczos, and
  Salakhutdinov]{li2018point}
Li, C.-L., Zaheer, M., Zhang, Y., Poczos, B., and Salakhutdinov, R.
\newblock Point cloud gan.
\newblock \emph{arXiv preprint arXiv:1810.05795}, 2018.

\bibitem[Makhzani et~al.(2015)Makhzani, Shlens, Jaitly, Goodfellow, and
  Frey]{makhzani2015adversarial}
Makhzani, A., Shlens, J., Jaitly, N., Goodfellow, I., and Frey, B.
\newblock Adversarial autoencoders.
\newblock \emph{arXiv preprint arXiv:1511.05644}, 2015.

\bibitem[Maturana \& Scherer(2015)Maturana and Scherer]{maturana2015voxnet}
Maturana, D. and Scherer, S.
\newblock Voxnet: A 3d convolutional neural network for real-time object
  recognition.
\newblock In \emph{2015 IEEE/RSJ International Conference on Intelligent Robots
  and Systems (IROS)}, pp.\  922--928. IEEE, 2015.

\bibitem[Qi et~al.(2017{\natexlab{a}})Qi, Su, Mo, and Guibas]{qi2017pointnet}
Qi, C.~R., Su, H., Mo, K., and Guibas, L.~J.
\newblock Pointnet: Deep learning on point sets for 3d classification and
  segmentation.
\newblock In \emph{Proceedings of the IEEE Conference on Computer Vision and
  Pattern Recognition}, pp.\  652--660, 2017{\natexlab{a}}.

\bibitem[Qi et~al.(2017{\natexlab{b}})Qi, Yi, Su, and Guibas]{qi2017pointnet++}
Qi, C.~R., Yi, L., Su, H., and Guibas, L.~J.
\newblock Pointnet++: Deep hierarchical feature learning on point sets in a
  metric space.
\newblock In \emph{Advances in neural information processing systems}, pp.\
  5099--5108, 2017{\natexlab{b}}.

\bibitem[Rubner et~al.(2000)Rubner, Tomasi, and Guibas]{rubner2000earth}
Rubner, Y., Tomasi, C., and Guibas, L.~J.
\newblock The earth mover's distance as a metric for image retrieval.
\newblock \emph{International journal of computer vision}, 40\penalty0
  (2):\penalty0 99--121, 2000.

\bibitem[Sheikh et~al.(2017)Sheikh, Rasul, Merentitis, and
  Bergmann]{sheikh2017stochastic}
Sheikh, A.-S., Rasul, K., Merentitis, A., and Bergmann, U.
\newblock Stochastic maximum likelihood optimization via hypernetworks.
\newblock \emph{arXiv preprint arXiv:1712.01141}, 2017.

\bibitem[Shoef et~al.(2019)Shoef, Fogel, and Cohen-Or]{shoef2019pointwise}
Shoef, M., Fogel, S., and Cohen-Or, D.
\newblock Pointwise: An unsupervised point-wise feature learning network.
\newblock \emph{arXiv preprint arXiv:1901.04544}, 2019.

\bibitem[Stypu{\l}kowski et~al.(2019)Stypu{\l}kowski, Zamorski, Zi{\k{e}}ba,
  and Chorowski]{stypulkowski2019conditional}
Stypu{\l}kowski, M., Zamorski, M., Zi{\k{e}}ba, M., and Chorowski, J.
\newblock Conditional invertible flow for point cloud generation.
\newblock \emph{arXiv preprint arXiv:1910.07344}, 2019.

\bibitem[Su et~al.(2015)Su, Maji, Kalogerakis, and Learned-Miller]{su2015multi}
Su, H., Maji, S., Kalogerakis, E., and Learned-Miller, E.
\newblock Multi-view convolutional neural networks for 3d shape recognition.
\newblock In \emph{Proceedings of the IEEE international conference on computer
  vision}, pp.\  945--953, 2015.

\bibitem[Sun et~al.(2018)Sun, Wang, Liu, Siegel, and Sarma]{sun2018pointgrow}
Sun, Y., Wang, Y., Liu, Z., Siegel, J.~E., and Sarma, S.~E.
\newblock Pointgrow: Autoregressively learned point cloud generation with
  self-attention.
\newblock \emph{arXiv preprint arXiv:1810.05591}, 2018.

\bibitem[Tomczak \& Welling(2017)Tomczak and Welling]{tomczak2017vae}
Tomczak, J.~M. and Welling, M.
\newblock Vae with a vampprior.
\newblock \emph{arXiv preprint arXiv:1705.07120}, 2017.

\bibitem[Tran(2013)]{tran20133d}
Tran, M.-P.
\newblock 3d contour closing: A local operator based on chamfer distance
  transformation.
\newblock 2013.

\bibitem[Van~den Oord et~al.(2016)Van~den Oord, Kalchbrenner, Espeholt,
  Vinyals, Graves, et~al.]{van2016conditional}
Van~den Oord, A., Kalchbrenner, N., Espeholt, L., Vinyals, O., Graves, A.,
  et~al.
\newblock Conditional image generation with pixelcnn decoders.
\newblock In \emph{Advances in neural information processing systems}, pp.\
  4790--4798, 2016.

\bibitem[Wu et~al.(2016)Wu, Zhang, Xue, Freeman, and Tenenbaum]{wu2016learning}
Wu, J., Zhang, C., Xue, T., Freeman, B., and Tenenbaum, J.
\newblock Learning a probabilistic latent space of object shapes via 3d
  generative-adversarial modeling.
\newblock In \emph{Advances in neural information processing systems}, pp.\
  82--90, 2016.

\bibitem[Wu et~al.(2015)Wu, Song, Khosla, Yu, Zhang, Tang, and Xiao]{wu20153d}
Wu, Z., Song, S., Khosla, A., Yu, F., Zhang, L., Tang, X., and Xiao, J.
\newblock 3d shapenets: A deep representation for volumetric shapes.
\newblock In \emph{Proceedings of the IEEE conference on computer vision and
  pattern recognition}, pp.\  1912--1920, 2015.

\bibitem[Yang et~al.(2018{\natexlab{a}})Yang, Luo, and Urtasun]{yang2018pixor}
Yang, B., Luo, W., and Urtasun, R.
\newblock Pixor: Real-time 3d object detection from point clouds.
\newblock In \emph{Proceedings of the IEEE conference on Computer Vision and
  Pattern Recognition}, pp.\  7652--7660, 2018{\natexlab{a}}.

\bibitem[Yang et~al.(2019)Yang, Huang, Hao, Liu, Belongie, and
  Hariharan]{yang2019pointflow}
Yang, G., Huang, X., Hao, Z., Liu, M.-Y., Belongie, S., and Hariharan, B.
\newblock Pointflow: 3d point cloud generation with continuous normalizing
  flows.
\newblock In \emph{Proceedings of the IEEE International Conference on Computer
  Vision}, pp.\  4541--4550, 2019.

\bibitem[Yang et~al.(2018{\natexlab{b}})Yang, Feng, Shen, and
  Tian]{yang2018foldingnet}
Yang, Y., Feng, C., Shen, Y., and Tian, D.
\newblock Foldingnet: Point cloud auto-encoder via deep grid deformation.
\newblock In \emph{Proceedings of the IEEE Conference on Computer Vision and
  Pattern Recognition}, pp.\  206--215, 2018{\natexlab{b}}.

\bibitem[Yifan et~al.(2019)Yifan, Wu, Huang, Cohen-Or, and
  Sorkine-Hornung]{yifan2019patch}
Yifan, W., Wu, S., Huang, H., Cohen-Or, D., and Sorkine-Hornung, O.
\newblock Patch-based progressive 3d point set upsampling.
\newblock In \emph{Proceedings of the IEEE Conference on Computer Vision and
  Pattern Recognition}, pp.\  5958--5967, 2019.

\bibitem[Yu et~al.(2018)Yu, Li, Fu, Cohen-Or, and Heng]{yu2018pu}
Yu, L., Li, X., Fu, C.-W., Cohen-Or, D., and Heng, P.-A.
\newblock Pu-net: Point cloud upsampling network.
\newblock In \emph{Proceedings of the IEEE Conference on Computer Vision and
  Pattern Recognition}, pp.\  2790--2799, 2018.

\bibitem[Zaheer et~al.(2017)Zaheer, Kottur, Ravanbakhsh, Poczos, Salakhutdinov,
  and Smola]{zaheer2017deep}
Zaheer, M., Kottur, S., Ravanbakhsh, S., Poczos, B., Salakhutdinov, R.~R., and
  Smola, A.~J.
\newblock Deep sets.
\newblock In \emph{Advances in neural information processing systems}, pp.\
  3391--3401, 2017.

\bibitem[Zamorski et~al.(2018)Zamorski, Zi{\k{e}}ba, Klukowski, Nowak, Kurach,
  Stokowiec, and Trzci{\'n}ski]{zamorski2018adversarial}
Zamorski, M., Zi{\k{e}}ba, M., Klukowski, P., Nowak, R., Kurach, K., Stokowiec,
  W., and Trzci{\'n}ski, T.
\newblock Adversarial autoencoders for compact representations of 3d point
  clouds.
\newblock \emph{arXiv preprint arXiv:1811.07605}, 2018.

\end{thebibliography}

\end{document}